\documentclass[letterpaper, 10 pt, conference]{ieeeconf}  

\IEEEoverridecommandlockouts                              

\usepackage{amsmath} 
\usepackage{amssymb}  
\usepackage{todonotes}
\usepackage{microtype}
\usepackage{url}
\usepackage{nicefrac}
\usepackage{graphicx}
\usepackage[caption=false]{subfig}  
\usepackage{tabularx}

\makeatletter
\newcommand\footnoteref[1]{\protected@xdef\@thefnmark{\ref{#1}}\@footnotemark}
\makeatother

\usepackage{tikz}
\usepackage{pgfplots}
\usetikzlibrary{arrows,decorations,backgrounds,shapes,chains}
\pgfplotsset{compat=newest}
\pgfplotsset{plot coordinates/math parser=false}
\newlength\figureheight
\newlength\figurewidth

\newcommand{\myvec}[1]{\boldsymbol{#1}}

\newcommand{\vf}{\myvec{f}}

\newcommand{\vp}{\myvec{p}}
\newcommand{\vq}{\myvec{q}}

\newcommand{\vu}{\myvec{u}}
\newcommand{\vv}{\myvec{v}}

\newcommand{\vx}{\myvec{x}}

\newcommand{\vC}{\myvec{C}}

\newcommand{\vG}{\myvec{G}}

\newcommand{\vJ}{\myvec{J}}

\newcommand{\vL}{\myvec{L}}
\newcommand{\vM}{\myvec{M}}

\newcommand{\vS}{\myvec{S}}
\newcommand{\vT}{\myvec{T}}

\title{\LARGE \bf Automatic Differentiation of Rigid Body Dynamics for Optimal Control and Estimation}

\author{
Markus Giftthaler\textsuperscript{a}$^{\dagger \ast}$,
Michael Neunert\textsuperscript{a}$^\dagger$,
Markus St\"auble\textsuperscript{a},
Marco Frigerio\textsuperscript{b},
Claudio Semini\textsuperscript{b} 
and Jonas Buchli\textsuperscript{a}
\thanks{
$^\dagger$These authors contributed equally to this work. \newline
$^\ast$Corresponding author. Email: mgiftthaler@ethz.ch \newline
\textsuperscript{a}Agile \& Dexterous Robotics Lab, Institute of Robotics and Intelligent Systems, ETH Z\"urich, Switzerland; \newline
\textsuperscript{b}Dept. of Advanced Robotics, Istituto Italiano di Tecnologia, Genova, Italy}
}%

\begin{document}
\maketitle
\thispagestyle{empty}
\pagestyle{empty}

\begin{abstract}
Many algorithms for control, optimization and estimation in robotics depend on derivatives of the underlying system dynamics, e.g. to compute linearizations, sensitivities or gradient directions. However, we show that when dealing with Rigid Body Dynamics, these derivatives are difficult to derive analytically and to implement efficiently. To overcome this issue, we extend the modelling tool `RobCoGen' to be compatible with Automatic Differentiation. Additionally, we propose how to automatically obtain the derivatives and generate highly efficient source code. 
We highlight the flexibility and performance of the approach in two application examples. First, we show a Trajectory Optimization example for the quadrupedal robot HyQ, which employs auto-differentiation on the dynamics including a contact model. 
Second, we present a hardware experiment in which a 6 DoF robotic arm avoids a randomly moving obstacle in a go-to task by fast, dynamic replanning. This paper is an extended version of~\cite{neunert:2016:derivatives}.
\end{abstract}


\section{Introduction}
Most robotic systems consist of multiple rigid links connected via joints. This includes entire systems such as robotic arms, legged robots or exoskeletons, but also components like robotic hands or integrated actuators. These systems can be mathematically modeled using Rigid Body Dynamics (RBD). The latter results in non-linear differential equations for which computing a closed-form solution is intractable. Therefore, many state-of-the-art approaches in control, estimation, optimization and planning rely on iterative, gradient-based algorithms which employ linear approximations of the given system equations. These algorithms include optimal controllers such as Linear Quadratic Regulators (LQR), optimal estimation approaches such as Kalman Filters and Batch Estimation, as well as parametric design optimization and Trajectory Optimization (TO), e.g. Direct Transcription,  Direct Multiple Shooting and Differential Dynamic Programming.

To obtain the required derivatives for these algorithms, there are several options. Often, analytic expressions `manually-derived on paper' are considered the ideal solution since they are accurate and fast to compute. However, as shown later, the expressions obtained from deriving RBD derivatives manually are fairly complex and difficult to optimize. Therefore, they often lead to poor runtimes. Additionally, the manual process is error prone. To avoid this step, symbolic toolboxes such as Matlab~\cite{web:matlab}, Mathematica~\cite{web:mathematica} or Maxima~\cite{web:maxima} can be employed. These tools apply known calculus rules in order to symbolically determine the derivatives. While this approach is viable in theory, in practice the derivative expressions easily get too large to be suitable for a computationally efficient implementation.

As a shortcut to the previously mentioned approaches, Numeric Differentiation (Num-Diff) is frequently used. In this method, the input to the function to be differentiated is perturbed in each input dimension to obtain an approximation of the derivative using finite differences. However, these methods are prone to numeric errors and they are computationally costly when the input dimension is high.

As another option, Automatic Differentiation, which is often also referred to as `Algorithmic Differentiation', can be used. Automatic Differentiation (Auto-Diff) is a programming approach for obtaining the derivatives from source code instructions. Auto-Diff builds an expression graph of the original function which is later differentiated using the chain rule and known `atomic' derivatives. Since Auto-Diff operates on expressions rather than numerical values, it provides the same accuracy as analytical derivatives. Yet, it still offers the comfort of obtaining the derivatives in an automated fashion. Additionally, Auto-Diff derivatives usually have lower complexity than unoptimized analytical derivatives.

\subsection{Automatic Differentiation Tools}
Over the years, several Automatic Differentiation tools have been developed. One implementation approach is Source Code Transform, i.e. parsing the (uncompiled) source code to build the expression graph and subsequently generating code for its derivatives. 
However, for advanced programming languages such as C++, this is challenging to implement. Hence, where supported by the programming language, Auto-Diff tools often rely on operator overloading. In this technique, the function to be differentiated is called with a special Auto-Diff scalar type instead of standard numeric types. This scalar type has overloaded operators that `record' the operations performed on it. The Auto-Diff tool then builds the expression graph from the recordings. Popular libraries in C++ that employ operator overloading are e.g. Adept~\cite{hogan:2014:adept}, Adol-C~\cite{walther:2012:adolc}, CppAD~\cite{bell:2012:cppad} and FADBAD++~\cite{bendtsen:1996:fadbad}. While all these tools are based on the same approach, there are significant differences both in performance as well as in functionality~\cite{carpenter:2015:stan, hogan:2014:adept}. The performance difference usually stems from the implementation of the expression graph, i.e. how fast it can be differentiated and evaluated. In terms of functionality, especially higher order derivatives make a key difference. They are difficult to implement and thus not supported by many tools. 

In this work, we are not proposing a new Auto-Diff tool but rather discuss how the existing tools can be leveraged when working with Rigid Body Dynamics and underline their benefits for efficient robotics control and optimization.

\subsection{Contributions}
Auto-Diff is already widely used in the mathematical optimization community - but it is only slowly gaining popularity in the robotics community. In this work, we illustrate the potential of using Auto-Diff for robotics. We demonstrate that it outperforms analytical derivatives in terms of computational complexity while preventing significant overhead in obtaining them. To employ Auto-Diff on Rigid Body Dynamics, we extend our open source modelling tool `RobCoGen'~\cite{frigerio:2016:robcogen} to be Auto-Diff compatible. The resulting framework allows for obtaining optimized derivatives of well-known RBD quantities and algorithms as used by most optimization, estimation and optimal control based algorithms.

Furthermore, rather than directly applying Auto-Diff at runtime, we perform an Auto-Diff code-generation step for the derivatives. In this step, the dynamic expression graph is converted to pure C~code. This eliminates the overhead of the expression graph, which we demonstrate is crucial for good performance. Furthermore, the resulting C~code is real-time capable and can be run on micro-controllers, in multi-threaded applications or even on GPUs. This allows for directly using it in hard real-time control loops and embedded platforms but also for massive sampling in data driven methods. Since RobCoGen is an open source library, we provide the community with a tested, efficient and easy to use tool for modelling, analyzing and controlling Rigid Body Systems.

To the best of the authors' knowledge, to date there is only one other robot control library~\cite{tedrake:2014:drake} employing rigid-body dynamics and partially supporting Automatic Differentiation. However, the tool chain presented in this work is the only RBD framework that has shown to also support derivative code generation.

This paper is an extended version of~\cite{neunert:2016:derivatives}. In this work, we provide additional timings and technical details on the implementation. Furthermore, we showcase two practical examples how our approach can lead to a performance leap in robotic control problems. 
First, we demonstrate how Auto-Diff can be used when modelling a Rigid Body System subject to contacts. We show that Auto-Diff can produce efficient derivatives of the system dynamics of a legged robot including a contact model. We demonstrate that this leads to a significant speed-up in optimal control applications.
The second example extends our previous example~\cite{neunert:2016:derivatives} on Auto-Diff for TO using Direct Multiple Shooting (DMS). While the original example showed a planning approach for collision-free trajectories for a robotic arm, we extend it to continuous online replanning. Robots often operate in uncertain, dynamically changing environments or in the presence of disturbances, requiring updates of the optimal control policy at runtime. We present a hardware experiment with a 6~DoF robot arm avoiding an unpredictably moving obstacle during a positioning task. Thanks to the low algorithm run-times achieved through the Auto-Diff generated derivative code, the replanning can be run in closed-loop with a robotic vision system.

\subsection{Related Work}
Existing libraries that implement auto differentiable Rigid Body Dynamics are rare. Drake~\cite{tedrake:2014:drake} supports Auto-Diff for gradient computations of dynamics. However, it relies on Eigen's~\cite{web:eigen} Auto-Diff module which cannot provide higher order derivatives and does not support code-generation for derivatives. Also, the code is not optimized for speed. Instead, it relies on dynamic data structures, which introduces significant overhead and limits the usability of the library for hard real-time, multi-threaded and embedded applications. MBSlib~\cite{wojtusch:2016:mbslib} also provides Auto-Diff support and does so in a more rigorous manner. Yet, it also relies on dynamic data structures, limiting efficiency and potential embedded and control applications. It is unclear whether MBSlib is compatible with any existing Auto-Diff code-generation framework.

There are also efforts of deriving simplified analytic derivatives of RBD~\cite{garofalo:2013:derivatives}. However, the resulting expressions have to be implemented manually and do not necessarily fully exploit efficient RBD algorithms, leading to more complex expressions with runtime overhead. A thorough discussion on this issue is presented in Section~\ref{sec:rbd}. A comparison between Symbolic and Automatic Differentiation for Rigid Body Kinematics is conducted in~\cite{durrbaum:2002:comparison}. However, the authors do not perform a code-generation step for Auto-Diff which, as we will see later, significantly improves performance.

There is considerable research on how to use Auto-Diff to model and simulate Rigid Body systems, e.g. \cite{eberhard1999automatic}, \cite{griffith:2005:some}, \cite{callejo:2014:performance}. However, this work is focused on solving the RBD differential equations in order to obtain equations of motions rather than explicitly computing their derivatives. Similar approaches are also used to perform a sensitivity analysis or parametric design optimization, as e.g. in \cite{bischof1996automatic}. However, this research field is only partially concerned with explicit derivatives of kinematics and dynamics algorithms. Additionally, metrics such as complexity and runtime, which are crucial for online control and estimation, are not the primary focus of this community.

In the context of our application examples, the optimal control tool chains `ACADO'~\cite{houska:2011:acado} and `CasADi'~\cite{andersson:2013:casadi} deserve special mentioning. The former allows users to specify optimal control problems with a symbolic syntax and solves them. The latter is a symbolic framework for automatic differentiation, which also supports code generation and can be interfaced with C code.  Both toolkits are targeted at real-time algorithms for control and optimization.
However, they lack the infrastructure to efficiently model Rigid Body systems and automatically compute relevant RBD quantities such as contact Jacobians, joint space inertia matrices, transforms or forward / inverse dynamics and kinematics. 
Instead of adding such an infrastructure to a general purpose tool, we decided to augment `RobCoGen', a proven tool for modelling RBD with Auto-Diff support. We then complement it with our own optimal control problem solver suite.

\section{Rigid Body Dynamics and Kinematics} 
\label{sec:rbd}
When dealing with Rigid Body Systems, we are usually interested in
forward/inverse dynamics, forward/inverse kinematics, the transforms between joints/links and end-effector/contact Jacobians. We express the Rigid Body Dynamics as
\begin{equation}
    \vM(\vq)\ddot{\vq} + \vC(\vq,\dot{\vq}) + \vG(\vq) = \vJ^\top_c(\vq) \lambda + \vS^\top \tau
    \label{eq:rbd}
\end{equation}
where $\vq$ represents the rigid body state in generalized coordinates, $\vM$ is the Joint Space Inertia matrix, $\vC$ are the Coriolis and centripetal terms, $\vG$ is the gravity term, $\vJ_c$ is the contact Jacobian, $\lambda$ are external forces/torques, $\vS$ is the selection matrix and $\tau$ represents the joint forces/torques. $\vS$ maps input forces/torques to joints and is used to model underactuated systems. In case of a fully actuated system with directly driven joints, $\vS$ is identity. For readability, we drop the dependency of these quantities on $\vq$, $\dot{\vq}$ in the following.

\subsection{Forward Dynamics and its Derivatives} 
\label{sub:forward_dynamics}
The forward dynamics equation
\begin{equation}
    fd(\vq, \dot{\vq}, \tau) = \ddot{\vq} = \vM^{-1} ( \vJ^\top_c \lambda + \vS^\top \tau - \vC - \vG )
    \label{eq:forward_dynamics}
\end{equation}
describes how a system reacts to given joint torques and external forces, in terms of generalized coordinates $\vq$.
Featherstone~\cite{roy:2008:rbda} shows that computing $\vM$ (without inverting it) has a worst-case complexity $\mathcal{O}(n^3)$ where $n$ is the number of rigid bodies, which makes naively evaluating Equation~\eqref{eq:forward_dynamics} very expensive. Therefore, he proposes several algorithms for factorizing $\vM$ and its inversion. Additionally, he proposes the Articulated Rigid Body Algorithm, which computes the forward dynamics, Equation~\eqref{eq:forward_dynamics}, with complexity $\mathcal{O}(n)$.

To get the derivatives of~\eqref{eq:forward_dynamics}, the chain rule can be applied:

\begin{align}
    &\frac{\partial (fd)}{\partial \vq} = \frac{d\vM^{-1}}{d\vq} \left( \vJ^\top_c \lambda + \vS^\top \tau - \vC - \vG \right) \label{eq:forward_dynamics_derivatives_q} \\
    &  ~~~ ~~~  ~~~              + \vM^{-1} \left(\frac{d\vJ^\top_c \lambda}{d\vq} - \frac{d\vC}{d\vq} - \frac{d\vG}{d\vq} \right) \nonumber \\
    &\frac{\partial (fd)}{\partial \dot\vq} = \vM^{-1} \frac{d\vC}{d\dot\vq}  \label{eq:forward_dynamics_derivatives_qd} \\
    &\frac{\partial (fd)}{\partial \tau} = \vM^{-1} \vS^\top
    \label{eq:forward_dynamics_derivatives_tau}
\end{align}
The expressions can be further simplified using the identities
\begin{align}
    & \frac{d\vM^{-1}}{d\vq} = \vM^{-1} \frac{d\vM}{d\vq} \vM^{-1} \\
    & \frac{d\vM}{d\vq} = \sum_{n=0}^N \frac{\partial \vJ^\top_n}{\partial \vq} \theta_n \vJ_n + \vJ^\top_n \theta_n  \frac{\partial \vJ^\top_n}{\partial \vq}
    \label{eq:identity_dmdq}
\end{align}
where $n$ is the index of a rigid body, $\theta_n$ is its fixed inertia matrix and $J_n$ is its state-dependent Jacobian. The identity in Equation~\eqref{eq:identity_dmdq} has been derived in~\cite{garofalo:2013:derivatives} and similar identities are presented for $\nicefrac{d\vC}{d\vq}$, $\nicefrac{d\vC}{d\dot\vq}$ and $\nicefrac{d\vG}{d\vq}$. But even without looking at these additional identities, two issues with analytical derivatives become prominent: First, the expressions are fairly large and error prone to implement. Second, there is significant amount of intermediate calculations in the forward dynamics that must be handled carefully to avoid re-computation. Thus, there are three tedious, manual steps involved in implementing analytical derivatives: (i) Careful derivation of the formulas, (ii) their correct implementation and (iii) intelligent caching of intermediate results. As we show in the experiments, Auto-Diff takes care of all three steps, alleviating the user from all manual work while still providing an equally fast or even faster implementation.

\subsection{Inverse Dynamics and its Derivatives} 
\label{sub:inverse_dynamics}
If we want to know what joint torques are required to achieve a certain acceleration at a certain state, we can compute the inverse dynamics by solving Equation~\eqref{eq:rbd} for $\tau$. When actuating all joints directly, $\vS$ is identity and we get
\begin{equation}
    id(\vq, \dot{\vq}, \ddot{\vq}) = \tau = \vM\ddot{\vq} + \vC + \vG - \vJ^\top_c \lambda \ \text{.}
    \label{eq:inverse_dynamics}
\end{equation}
For the inverse dynamics computation, Featherstone presents an efficient algorithm, the Recursive Newton-Euler Algorithm~\cite{roy:2008:rbda}, which again has complexity $\mathcal{O}(n)$ and avoids computing $\vM$ and other elements explicitly. 
For the fully actuated case, the inverse dynamics derivatives are defined as
\begin{align}
    &\frac{\partial (id)}{\partial \vq} = \frac{d\vM}{d\vq}\ddot{\vq} +   \frac{d\vC}{d\vq} + \frac{d\vG}{d\vq} - \frac{d\vJ^\top_c}{d\vq} \lambda \label{eq:inverse_dynamics_derivatives_1}\\
    &\frac{\partial id}{\partial \dot\vq} = \frac{d\vC}{d\dot\vq} \label{eq:inverse_dynamics_derivatives_2} \\
    &\frac{\partial id}{\partial \ddot{\vq}} = \vM
    \label{eq:inverse_dynamics_derivatives_3}
\end{align}
If we were to analytically simplify these equations further, we could again use the identity in Equation~\eqref{eq:identity_dmdq} and other identities presented in~\cite{garofalo:2013:derivatives}.

In case of an under-actuated robot, such as floating base robots, we can obtain the inverse dynamics by using an (inertia-weighted) pseudo-inverse of $\vS$ to solve Equation~\eqref{eq:rbd} for $\tau$ as described in~\cite{nakanishi:2007:inverse}. This is only one possible choice for computing floating base inverse dynamics~\cite{righetti:2011:inverse}. However, it will serve as an example and the following discussion extends to the other choices as well.
The inertia-weighted pseudo-inverse is defined as 
\begin{equation*}
{}\bar{\vS} = \vM^{-1} \vS^\top (\vS \vM^{-1} \vS^\top)^{-1}
\end{equation*}
and leads to the inverse dynamics expression
\begin{align}
    &id(\vq, \dot{\vq}, \ddot{\vq}) = \tau \notag \\
    &= (\vS \vM^{-1} \vS^\top)^{-1} \ddot{\vq} + \bar{\vS}^\top (\vC + \vG - \vJ^\top_c \lambda)
    \label{eq:inverse_dynamics_pseudoinverse} \quad \text{.}
\end{align}
The derivatives of the inverse dynamics then become
\begin{align}
    &\frac{\partial (id)}{\partial \vq} = \frac{d (\vS \vM^{-1} \vS^\top)^{-1}}{d\vq}\ddot{\vq} +   \frac{d\bar{\vS}^\top}{d\vq} (\vC + \vG - \vJ^\top_c \lambda) \notag \\
    & ~~~ ~~~ ~~~ + \bar{\vS}^\top\left( \frac{d\vC}{d\vq} + \frac{d\vG}{d\vq} - \frac{d\vJ^\top_c}{d\vq}\right)
    \label{eq:inverse_dynamics_derivatives_pseudo_1}\\
    &\frac{\partial (id)}{\partial \dot\vq} = \bar{\vS}^\top \frac{d\vC}{d\dot\vq} \label{eq:inverse_dynamics_derivatives_pseudo_2} \\
    &\frac{\partial (id)}{\partial \ddot{\vq}} = (\vS \vM^{-1} \vS^\top)^{-1} \quad \text{.}
    \label{eq:inverse_dynamics_derivatives_pseudo_3}
\end{align}
Similar to the forward dynamics case, we see that the derivative expressions become large, and their implementation is similarly error prone. Additionally, without careful optimization, we might accidentally introduce significant overhead in the computation, e.g. by not caching intermediate results.

\subsection{Higher-Order Derivatives}
In Subsections~\ref{sub:forward_dynamics} and~\ref{sub:inverse_dynamics} we presented the first-order derivatives of the forward and inverse dynamics. However, for some optimal control algorithms such as Differential Dynamic Programming (DDP)~\cite{mayne1966ddp}, we need second order derivatives of the system dynamics. If we want to further analytically differentiate the derivatives, we have to apply the chain rule to them, resulting in even larger expressions. At this point it becomes highly questionable if analytic derivatives should be implemented manually. Since symbolic computation engines also apply the chain rule and only have limited simplification capabilities, these approaches will face the same issues for second order derivatives. This is also part of the reason why second order optimal control algorithms, such as DDP, are only rarely applied to more complex robotic systems. Instead, algorithms that approximate the second order derivatives are usually preferred, see for example~\cite{slq}. When using Auto-Diff, we can easily obtain second order derivatives and hence do not face this limitation.

\section{Automatic Differentiation with RobCoGen}
\subsection{The Robotics Code Generator}
The Robotics Code Generator (RobCoGen) is a computer program that, given the description of an articulated robot, generates optimized code for its kinematics and dynamics~\cite{frigerio:2016:robcogen}. A simple file format is available to provide the description (model) of the robot. Currently, RobCoGen generates both C++ and Matlab code, implements coordinate transforms, geometric Jacobians and state-of-the-art algorithms for
forward and inverse dynamics, as described in~\cite{roy:2008:rbda}.

\subsection{Derivatives for RobCoGen}
RobCoGen does not natively support any code generation for derivatives of rigid body kinematics or dynamics. However, it can be easily extended to generate C++ code suitable for Automatic Differentiation in order to leverage other tools specifically designed for that purpose,
such as CppAD. The changes required to support Auto-Diff essentially reduce to generating code templated on the scalar type (rather than using standard \texttt{float} or
\texttt{double}). This simple change enables the use of a variety of Auto-Diff tools based on operator overloading, yet does not prevent the regular usage of the generated code.

It is important to note that RobCoGen uses code-generation to create robot specific RBD code. This is not to be confused with Auto-Diff codegen, which uses RBD code to compute a derivative function, which is then translated into source code. In this work, we apply Auto-Diff codegen to the output of RobCoGen. Thus, there are two sequential, entirely unrelated code-generation steps involved in this work. First, RobCoGen generates the dynamics and kinematics functions. Then Auto-Diff codegen uses those to create the derivative source code. We will always indicate which type of generated code we are referring to, i.e. whether we refer to the dynamics/kinematics generated by RobCoGen or to the derivatives generated using Auto-Diff codegen\footnote{While the implementation of both code generation steps differs significantly, the goal is the same. In both cases, specialized code is generated instead of using general dynamic data structures. This eliminates the overhead of traversing such data structures, checking for dimensions, sizes and branches. Furthermore, static data structures are easier to implement on embedded systems and to use in hard real-time applications.}.

Since RobCoGen implements the most efficient algorithms for dynamics (e.g. the Articulated Rigid Body algorithm for forward dynamics), and since the lowest achievable complexity of Auto-Diff derivatives is proportional to the complexity of the original function \cite{griewank:2008:derivatives}, it follows that, at least theoretically, we can obtain the most efficient derivatives from RobCoGen as well.

\subsection{Implementation in RobCoGen}
Since the interoperability with Auto-Diff tools is ongoing development within the RobCoGen
project, we implemented a proof of concept of the approach by modifying existing RobCoGen generated code for the quadruped robot HyQ \cite{semini:2011:hyqjournal} and a 6~DoF robotic arm.

Our goal is to make the entire RobCoGen generated code auto-differentiable. Thus, we add a scalar template to all algorithms which compute one of the following quantities:
\begin{itemize}
\item Forward dynamics: Articulated Rigid Body Algorithm
\item Inverse dynamics: Recursive Newton-Euler Algorithm
\item Homogeneous-coordinate transforms
\item Coordinate transform for spatial vectors
\item Jacobians
\item Rigid Body quantities such as $\vM$, $\vC$ and $\vG$ 
\end{itemize}
We further template all input and parameter types, including state, joint torques, external forces, all inertia parameters etc. Therefore, also uncommon derivatives, e.g. derivatives of inverse dynamics with respect to inertia parameters, can be obtained. This feature can be useful for design optimization or parameter estimation applications. Also, Auto-Diff derivatives of essential Rigid Body quantities such as the Joint Space Inertia Matrix $\vM$ can be computed. This allows to use Auto-Diff to generate custom derivatives for special applications, e.g. when using Pseudo-Inverse based inverse dynamics as shown in Subsection~\ref{sub:inverse_dynamics}. In such a case, the derivatives in Equations~\eqref{eq:inverse_dynamics_derivatives_pseudo_1}~-~\eqref{eq:inverse_dynamics_derivatives_pseudo_3}, can still be obtained from Equation \eqref{eq:inverse_dynamics_pseudoinverse} using Auto-Diff. The user only has to ensure that the computation of the Pseudo-Inverse is auto-differentiable as well. Furthermore, the user can also generate derivatives of transforms and Jacobians, e.g. for task space control or kinematic planning.

By templating the entire library, the user can specify which methods or quantities to differentiate.  Therefore, the differentiation can be tailored to a specific use case, e.g. by computing only required parts of a Jacobian or applying the `Cheap Gradient Principle' \cite{griewank:2008:derivatives}. Another benefit is that not all derivatives are generated but only the ones required. To demonstrate how to select and generate derivatives, we provide the reference implementation described below.

\subsection{Reference Implementation of Automatic Differentiation}
In order to validate the performance and accuracy of the Auto-Diff compatible version of RobCoGen, we provide a reference implementation. For this implementation, we use CppAD as our Auto-Diff tool. CppAD is very mature, well documented, supports higher derivatives and provides an efficient implementation. Still, evaluating the expression graph for computing derivatives comes at an overhead. Therefore, we also employ CppADCodeGen~\cite{web:cppadcodegen}, which can generate bare C code for derivatives using CppAD.

Since CppAD operates on scalars rather than full matrices, the generated derivative code is only using scalar expressions and cannot leverage advanced compiler optimization such as vectorization. While this leads to a slight decrease in performance, the generated code is dependency-free and can easily be run on embedded systems or GPUs. Additionally, memory for intermediate results of the derivative computation can be allocated statically. Thus, the generated code can be run in hard real-time control loops. Despite the fact that the generated code is static, it can be parametrized. Therefore, RBD parameters such as mass, inertia or center of mass can be changed at runtime if required. Thus, the generated code is specialized to a certain morphology rather than a specific robot or a set of dynamic properties.

\section{Results}
\label{sec:results}
To evaluate the performance of our Auto-Diff approach, we first run standalone synthetic tests on the derivatives in terms of complexity and accuracy. These tests are performed on a model of the underactuated quadruped HyQ~\cite{semini:2011:hyqjournal}, which has a floating base and three joints per leg. We then present two application examples. The first example demonstrates Auto-Diff for motion planning on HyQ. The second one demonstrates online TO applied to a 6 degree of freedom arm and includes hardware experiments.

\subsection{Accuracy, Number of Instructions and Timings}
\label{subsec:timings}
In order to compare the different derivative methods, we look at the derivative of the forward dynamics with respect to the joint forces/torques as shown in Equation~\eqref{eq:forward_dynamics_derivatives_tau}. To ensure that we are computing a dense derivative, we ignore the floating base part of the equation but focus on the bottom right corner of the expression. For the comparison, the derivatives are obtained from different implementations: Single-sided numerical derivatives, Auto-Diff at runtime, Auto-Diff codegen and analytical derivatives. For the analytical derivatives we have two implementations. One computes $\vM$ using Featherstone's Composite Rigid Body algorithm and later naively inverts it by using Eigen's general $\vL\vL^\top$ solver. The second analytical implementation uses the RBD-specific $\vL^\top\vL$ factorization of $\vM^{-1}$, which exploits sparsity. Such a factorization is described in~\cite{roy:2008:rbda} and implemented by RobCoGen.

We first verify the accuracy by measuring the norm of the difference between the derivative matrices. As expected, the analytical methods and Auto-Diff agree up to machine precision ($<10^{-13}$). Numerical differentiation however deviates by around $<10^{-6}$ from all other implementations.

\begin{figure}[tbp]
    \centering
    \includegraphics[width = 0.99\columnwidth]{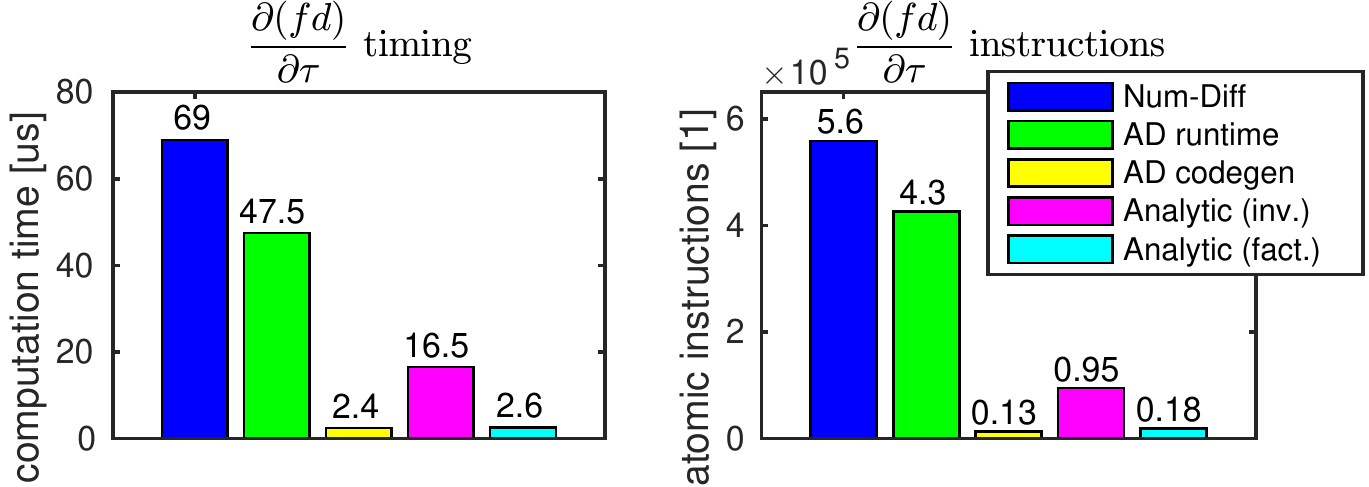}
    \caption{Comparison between different approaches to compute the forward dynamics derivative with respect to joint torques~$\tau$. We measure absolute computation time as well as the number of atomic instructions of the compiled code. Clearly, Num-Diff performs worst. Auto-Diff at runtime performs slightly better. However, Auto-Diff codegen and the analytic factorization perform best. The naive analytic inverse performs significantly worse than both. This underlines that naively implemented analytical derivatives can be significantly inferior to Auto-Diff.}
    \label{fig:raw_timings}
\end{figure}

Additionally, we compare the runtime as well as the number of atomic instructions of all derivative methods. In this test, we average over 10,000 computations and enable the highest optimization level of our compiler. 
The results of this test, run on an Intel Core i7-4850HQ CPU (2.30~GHz) are shown in Figure~\ref{fig:raw_timings}. This test shows two interesting findings: Firstly, there is a very significant difference between the Auto-Diff generated derivative code and the Auto-Diff computation at runtime. The generated code is much faster, which is possibly a result of removing the overhead of travelling through the expression graph and enabling the compiler to optimize the code. Secondly, we see that -- despite the fact that the RobCoGen's factorization of $\vM^{-1}$ exploits the sparsity and structure of the problem -- it is not able to outperform the Auto-Diff generated derivative in terms of instructions and runtime. The factorization is still about 10\% slower than Auto-Diff codegen. This strongly underlines the hypothesis that even carefully implemented analytical derivatives are usually equally or more expensive than those generated by Auto-Diff. On the other hand, naively implemented derivatives, even though analytic, can lead to significant overhead. This overhead, as present in the analytical inversion, is over 600\% in the test above.

\begin{figure}[tbp]
    \centering
    \includegraphics[width = 0.99\columnwidth]{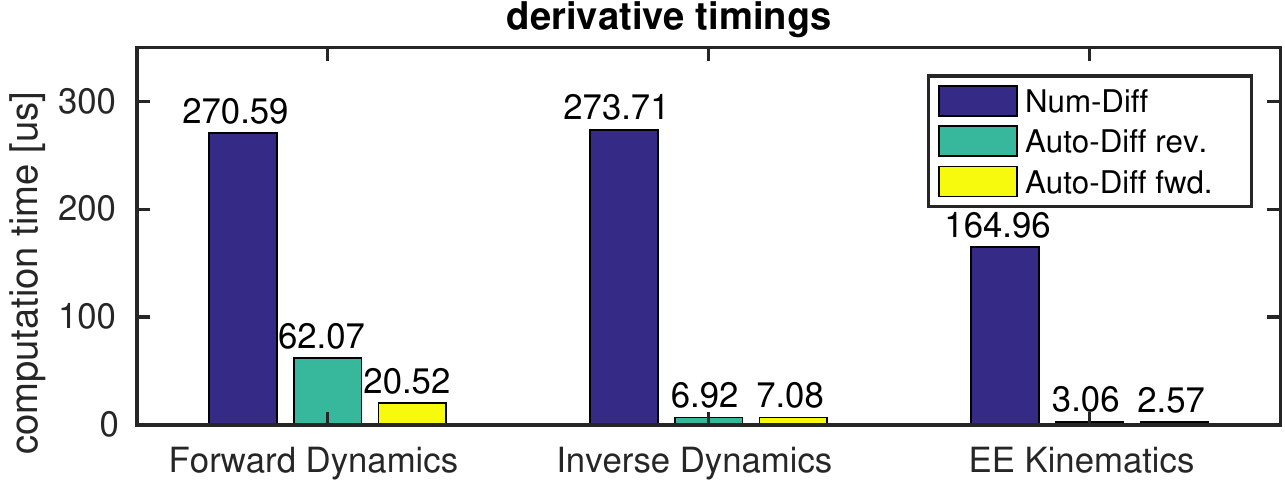}
    \caption{Comparison between Numeric Differentiation ('Num-Diff') and Automatic Differentiation codegen ('Auto-Diff') using reverse ('rev') and forward ('fwd') mode for the computation of three types of derivatives on HyQ. For the forward dynamics derivatives we implement Equations~\eqref{eq:forward_dynamics_derivatives_q}-\eqref{eq:forward_dynamics_derivatives_tau} while in the inverse dynamics derivatives we use Equations~\eqref{eq:inverse_dynamics_derivatives_1}-\eqref{eq:inverse_dynamics_derivatives_3}. Additionally, end-effector position and velocity given in Equation~\eqref{eq:forward_kin} differentiated with respect to the rigid body state $\vq$ and $\dot{\vq}$ are timed. We can see that Auto-Diff codegen outperforms Num-Diff by a factor of about 10-40x.}
    \label{fig:other_timings}
\end{figure}

Finally, we test more complex expressions. We compare the derivatives of the forward dynamics as well as of the fully-actuated inverse dynamics both taken w.r.t. the state, as shown in  Equations~\eqref{eq:forward_dynamics_derivatives_q}-\eqref{eq:forward_dynamics_derivatives_tau} and Equations~\eqref{eq:inverse_dynamics_derivatives_1}-\eqref{eq:inverse_dynamics_derivatives_3} respectively. Also here, we are not using the naive implementations but efficient Featherstone algorithms for all derivative approaches. Additionally, we time the computation of the end-effector position $\vp$ and velocity $\dot{\vp}$ differentiated with respect to the full state, $\vq$ and $\dot{\vq}$. Here, $\vp$ and $\dot{\vp}$ are expressed in a fixed inertial frame (`world' frame). The forward kinematics to be differentiated are given by 
\begin{align}
    \vp(\vq) = \vT_{ee}(\vq) \vq, ~ 
    \dot{\vp}(\vq, \dot{\vq}) = \vJ_{ee}(\vq, \dot{\vq}) \dot{\vq}
    \label{eq:forward_kin}
\end{align}
where $\vT_{ee}$ and $\vJ_{ee}$ are end-effector transform and Jacobian respectively. We compute the forward kinematics derivatives for all four feet of the quadruped. Since we do not have optimized analytical derivatives available for this test, we compare the performance of Num-Diff to the one of Auto-Diff codegen. In Auto-Differentiation, there are two internal modes of building the expression graph: Forward and reverse. While both lead to the same derivatives with the same accuracy, the graph structure might be simpler in one than in the other. Therefore, we include timings for both modes.

Results of this test are shown in Figure~\ref{fig:other_timings} which plots the average of 10,000 executions of each approach. There are several interesting observations to be made. First, Auto-Diff codegen is about 10-40x faster than Num-Diff. Secondly, Num-Diff timings for forward dynamics and inverse dynamics are of similar magnitude due to the fact that the derivatives are of similar dimensions and similarly complex to compute. However, Auto-Diff can generate a simpler structure for inverse dynamics than for forward dynamics, resulting in significantly shorter timings. Lastly, reverse mode is faster than forward mode for functions with significantly more input (`independent') than output (`dependent') variables. The ratios between input and output dimension on HyQ are 48:18 for the forward dynamics, 54:18 for the inverse dynamics and 36:24 for the forward kinematics of all legs. Despite the input dimension being larger than the output dimension for all three derivatives, we only see a slight benefit for the reverse mode in the inverse dynamics where the ratio is the largest. In the case of forward dynamics and kinematics, forward mode outperforms reverse mode. Since the input and output dimensions are fairly balanced, a general recommendation for which mode to use cannot be derived. However, the easy to remember rule of thumb ``forward mode for forward dynamics and forward kinematics, reverse mode for inverse dynamics'' can serve as an initial guess for performance optimization.

\subsection{Trajectory Optimization for a Quadrupedal Robot using Sequential Linear Quadratic Optimal Control}
\label{sec:hyq_slq}
To verify that the previously shown results make a difference in practical application, we are first looking at a TO problem for the quadrupedal robot HyQ. HyQ is a 18-DoF floating-base, underactuated robot, required to make and break contacts with the environment. Hence, planning motions for such a system is a high-dimensional problem. Such problems are often tackled using TO. Amongst others, TO problems can be transformed into Nonlinear Programs (NLPs), e.g. as shown in Subsection \ref{subsec:dms}, or formulated as DDP problems. In this example, we show how Auto-Diff codegen can improve performance of such a DDP approach, namely Sequential Linear Quadratic (SLQ) optimal control. 

\subsubsection{Sequential Linear Quadratic Optimal Control}
SLQ is an iterative optimal control algorithm which consists of three main steps. First, the system dynamics are forward integrated using the updated controller from a previous iteration or, at the first iteration, using a stable initial controller. The non-linear optimal control problem is then approximated by a linear quadratic optimal control problem by linearizing the system dynamics and quadratically approximating the cost around the forward simulated trajectory. Finally, the linear quadratic problem is solved in a backward pass using a Riccati approach. For a detailed description, we refer to~\cite{slq:2005}.

\subsubsection{SLQ Formulation for Legged Systems}
In our SLQ problem, we include the entire state of the robot
\begin{align}
    \vx &= [{}_W\vq^\top ~ {}_B\dot{\vq}^\top]^\top \notag 
    \\ &= [{}_W\Omega^\top_B ~ {}_W\vx^\top_B ~ \theta^\top ~ {}_B\omega^\top_B ~ {}_B\vv^\top_B ~ \dot{\theta}^\top]^\top
\end{align}
where ${}_W\Omega_B$ and ${}_W\vx_B$ define base orientation and position expressed in the inertial (`world') frame. ${}_B\omega_B$ and ${}_B\vv_B$ represent local angular velocity and linear acceleration expressed in a body fixed frame. Joint angles and velocities are represented by $\theta$ and $\dot{\theta}$, respectively.

The system dynamics are then defined as
\begin{align}
    \dot{\vx} &= [{}_W\dot{\vq}^\top ~ {}_B\ddot{\vq}^\top]^\top = \vf(\vx,\vu) \notag \\ 
    &= [H_{WB}({}_B\dot{\vq}) ~ fd(\vq, \dot{\vq},\vu)]^\top
    \label{eq:slq_sys}
\end{align}
where we set the control input equal to the joint torques $\vu = \tau$. 
$H_{WB}$ transforms velocities expressed in the body frame to the inertial frame. $fd(\cdot)$ represents the forward dynamics as shown in Equation~\eqref{eq:forward_dynamics}.

As mentioned before, HyQ is subject to contacts with the environment creating contact constraints and forces. Contacts can be included in SLQ either as constraints \cite{neunert:2016:humanoids} or using an explicit contact model \cite{neunert:2017:ral}. Here, we choose the latter approach. Assuming a static environment, the contact forces become a function of the current state of the robot, thus $\lambda = \lambda(\vq, \dot{\vq})$. In this example, we employ a `soft' contact model consisting of a non-linear spring in surface-normal direction and a non-linear damper for velocities. For each end-effector we compute the contact model in the contact frame~$C$ as follows:
\begin{align}
    {}_C\lambda(\vq, \dot{\vq}) = 
        &-k \exp(\alpha_k~{}_Cp_z(\vq)) \notag \\
        &-d~\text{sig}(\alpha_d~{}_Cp_z(\vq))~{}_C\dot{\vp}(\vq, \dot{\vq})
    \label{eq:contact_c}
\end{align}
where $k$ and $d$ define spring and damper constants. Since SLQ requires smooth derivatives \cite{neunert:2017:ral, neunert:2016:humanoids}, we multiply the damper with the sigmoid of the normal component $p_z(\vq)$ of the contact surface penetration $\vp(\vq)$ as defined in Equation~\eqref{eq:forward_kin}. While both the exponential and the sigmoid serve as smoothing functions, their `sharpness' is controlled with $\alpha_k$ and $\alpha_d$. Finally, the contact force is rotated into the body frame 
\begin{align}
    {}_B\lambda(\vq,\dot{\vq}) = R_{WB}(\vq) {}_C\lambda(\vq,\dot{\vq})
    \label{eq:contact_b}
\end{align}
before it is passed to the forward dynamics. When plugging Equations~\eqref{eq:contact_c}+\eqref{eq:contact_b} into the forward dynamics in Equation~\eqref{eq:forward_dynamics}, we can see that its derivatives in Equation~\eqref{eq:forward_dynamics_derivatives_q} become even more complex. Motivated by this, we apply Auto-Diff for linearizing the system dynamics~\eqref{eq:slq_sys} within SLQ.

\subsubsection{SLQ examples}
To test the performance influence of Auto-Diff codegen on SLQ, we define two test tasks. In each of the tasks, we provide a cost function to SLQ that penalizes the deviation from a desired final state and regularizes deviations from a nominal state. By varying the cost function parameters, we ask for forward motions of different length. In the first example, we request a forward motion of 1.5~m while in the second example we ask for 2~m. Since the time horizon is kept constant, task~1 leads to a trotting gait while task~2 leads to a galloping gait. These gaits are automatically discovered as described in \cite{neunert:2017:ral}. Both resulting motions are shown in the video attachment\footnote{https://youtu.be/rWmw-ERGyz4}.
All source code as well as the model, solver settings and cost function weights are available within our open source `Robotics and Optimal Control Toolbox'~\cite{neunert:2017:controltoolbox}.
\begin{figure}[tpb]
    \centering
    \subfloat[Learning rates for the SLQ example on HyQ. Both Num-Diff and Auto-Diff codegen exhibit the same rates when learning the same task. While not clearly visible, lines corresponding to the tasks overlap.]{
        \includegraphics[width=0.45\textwidth]{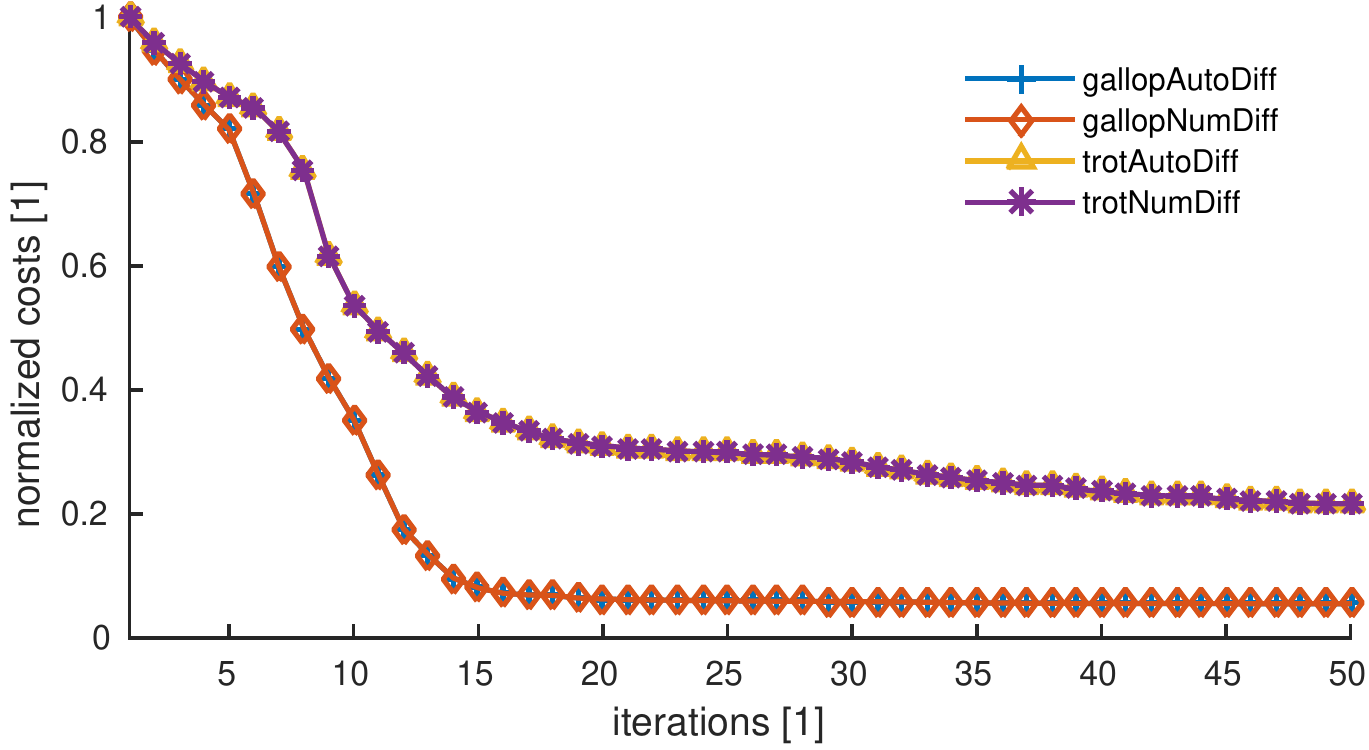}
        \label{fig:slq_convergence}
    } \qquad \hfill
    \subfloat[Runtime for SLQ iterations for different tasks and derivative types. Even though the linearization is only one part of the entire SLQ algorithm, using Auto-Diff codegen speeds up the overall algorithm by up to 500\%.]{
        \includegraphics[width=0.45\textwidth]{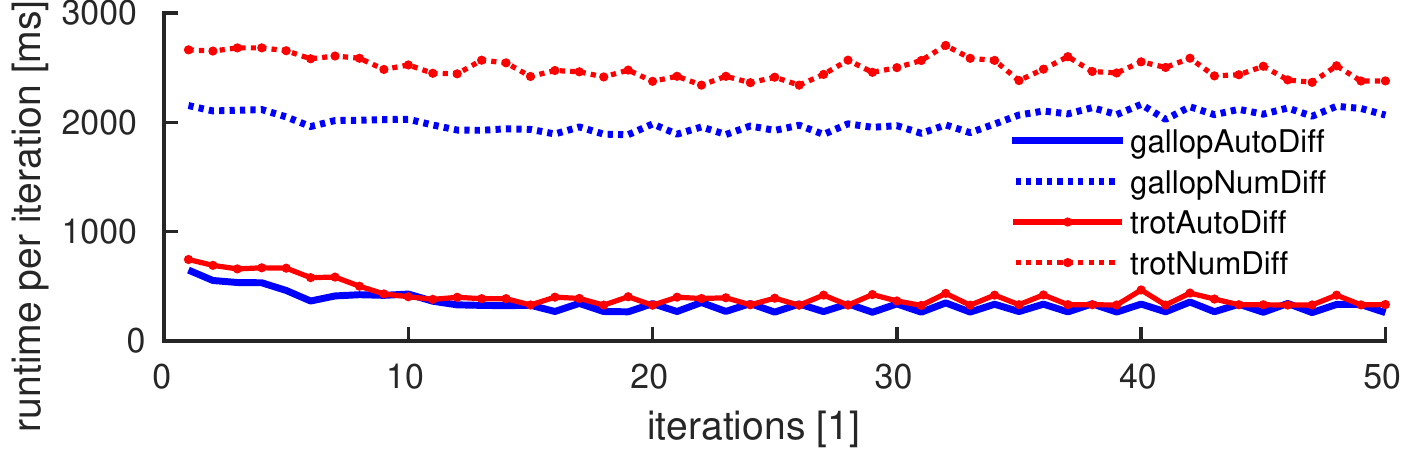}
        \label{fig:slq_runtime}
    }
    \caption{Speed and convergence tests for SLQ on two tasks for HyQ. All timings taken on an Intel Core i7 (2.8 GHz) CPU.}
    \label{fig:slq}
\end{figure}

For both tasks we compare SLQ with Auto-Diff codegen and with Num-Diff for the computation of the linear system approximation. Figure~\ref{fig:slq_convergence} shows that in both tasks Auto-Diff codegen and Num-Diff converge equally fast. While Num-Diff lacks precision, it does not impair convergence performance. However, when looking at the runtimes in Figure~\ref{fig:slq_runtime}, the advantage of Auto-Diff codegen becomes obvious. Auto-Diff codegen allows for a speedup of up to 500\% of the overall runtime. Note that system linearization is only one part of the SLQ algorithm. And while we do not achieve an overall speedup of around 10-40x as in the timing examples in Subsection~\ref{subsec:timings}, the runtime reduction is still significant. Given the fact that SLQ requires around 15 to 50 iterations until convergence for the given tasks, this results in an absolute runtime speedup of over a minute.

\subsection{Trajectory Optimization and fast trajectory replanning for a 6-DoF robot arm using Direct Multiple Shooting}
\label{subsec:dms}
In~\cite{neunert:2016:derivatives}, we presented a TO example with a fixed-base robot arm avoiding a static, spherical obstacle in a positioning task. Here, we extend this example as follows:
we assume that the obstacle can move in an unpredictable way -- to compensate for this, we perform dynamic trajectory replanning. 
The generated derivative code reduces the runtime of our algorithms significantly and allows us to achieve replanning frequencies that are high enough to run in closed-loop with a visual obstacle state estimator. 
We show hardware experiments on an industrial robot arm and demonstrate dynamic obstacle avoidance in a go-to task.

\subsubsection{Direct Multiple Shooting}
The Direct Multiple Shooting method~\cite{bock1984direct} is a widely-spread approach for numerically solving optimal control problems. An originally infinite-dimensional optimal control problem is transformed into an NLP by discretizing it into a finite set of state and control decision variables. Those are used for forward integrating the so-called `shots', which are matched at the nodes using continuity constraints. DMS can handle inequality path constraints, e.g. task space obstacles, or control input constraints. A detailed description of the method is beyond the scope of this paper, refer to~\cite{diehl2006fast} for an overview.

We have implemented a custom DMS problem generator which hands over the resulting Nonlinear Program to off-the-shelf NLP solvers such as IPOPT~\cite{wachter:2006:ipopt} and SNOPT~\cite{gill:2005:snopt}. Amongst others, it achieves efficiency through exploiting the inherent block-sparse structure of the DMS constraint Jacobian. We are using a standard fourth order Runge-Kutta (RK4) integration scheme for propagating the shots, and its analytic derivative for calculating the trajectory sensitivities w.r.t. neighbouring decision variables on the fly. The sensitivities are functions of the system dynamics derivatives w.r.t. state and control at every integration sub-step in time, which can be evaluated efficiently using the Auto-Diff generated derivatives.

\subsubsection{Robot go-to task with a randomly moving obstacle}
\begin{figure}[tbp]
    \centering
    \begin{tikzpicture}[scale=1]
    \node (pic) at (0,0) { \includegraphics[width =0.49\textwidth]{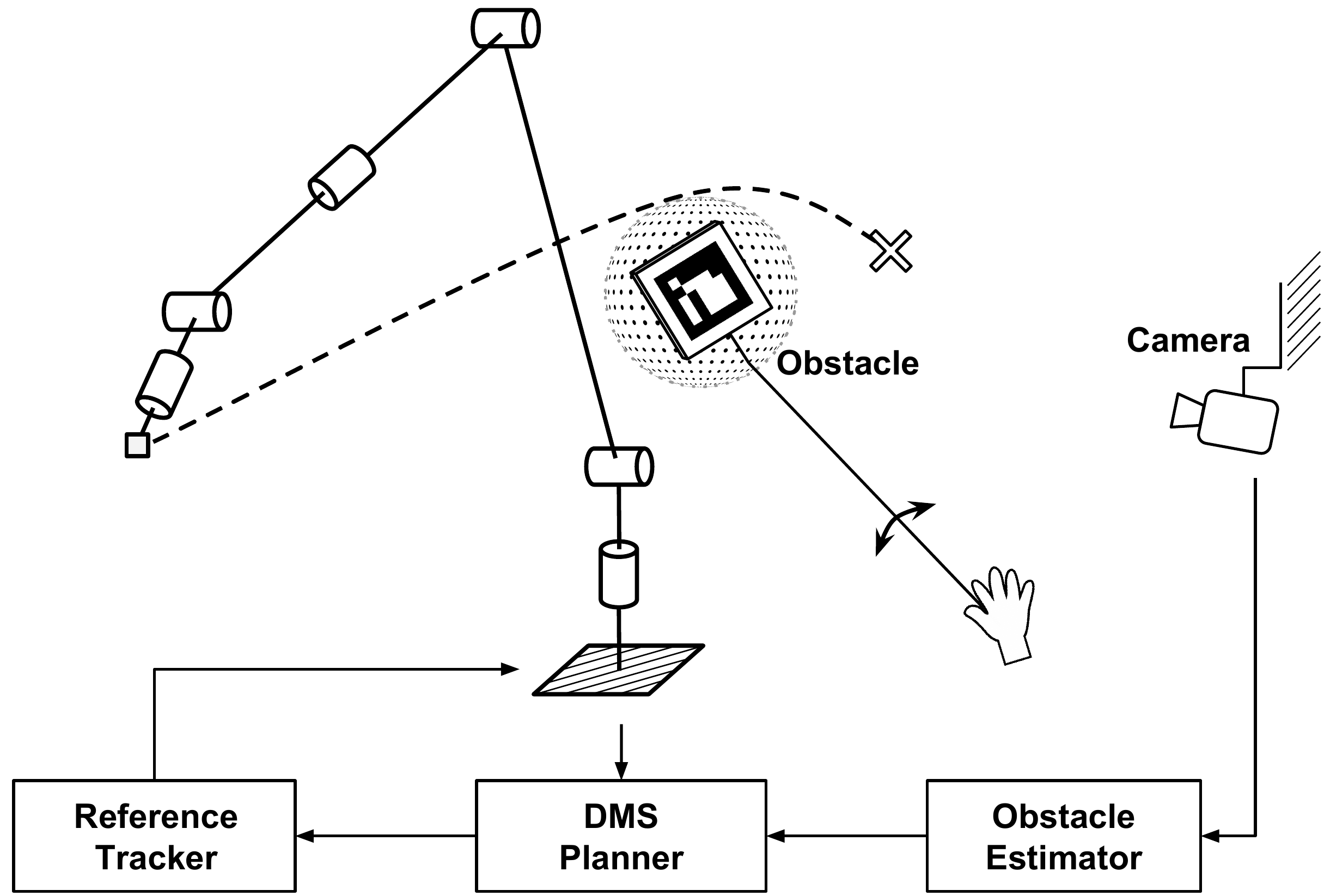}};
        \node [right] at (-3, -1.1){$\theta_{ref}, \dot \theta_{ref}$};
        \node [right] at (-0.2, -1.8){$\theta_m, \dot \theta_m$};
    \end{tikzpicture}
    \caption{The experimental setup: a fixed-base robot arm has to reach a desired joint configuration within a given time horizon~$T$ while avoiding a randomly moving obstacle.}
    \label{fig:abbExperiment}
\end{figure}
Figure~\ref{fig:abbExperiment} shows a sketch of the problem setup. A fixed-base robot arm has to reach a desired joint configuration within a given time horizon~$T$ while avoiding a randomly moving spherical task space obstacle.
In this experiment, we employ an ABB IRB4600 industrial robot arm with 2.55~m reach. The arm provides high-accuracy joint encoder readings ($\theta_m$), which allow to get reasonable joint velocity estimates ($\dot \theta_m$) by finite-differences. The arm is controlled through an interface which allows to set joint reference positions ($\theta_{ref}$) and velocities ($\dot \theta_{ref}$) at 250~Hz rate. Despite being bound to a kinematic controller, planning dynamically optimal trajectories is still an advantage in terms of energy efficiency~\cite{koen:2014:energy} and providing smooth, feasible references.

The moving obstacle in the robot's workspace is realized using a fiducial marker on a stick which is guided by a human. The obstacle is defined as a sphere with 0.5~m diameter around the marker's center. The marker is tracked by a camera system and an off-the-shelf visual state estimator provides position and velocity estimates for the obstacle at 20~Hz rate. The trajectory of the obstacle is predicted through forward integration of its current velocity estimate.
For obstacle collision avoidance, we define a grid of collision points distributed equally-spaced on the robot links. For each of those collision points, an inequality path constraint results at every node of the DMS problem.
The DMS planner accesses the latest robot and obstacle state estimates, and forwards its current optimal plan to a reference tracker which sets the joint state reference for the robot arm's joint controllers.

In order to deal with the changing obstacle, the planner needs to continuously re-compute optimal trajectories which drive the robot to the desired goal. In the literature, methods exist for nonlinear model predictive control or fast replanning using DMS, for example the approach presented in~\cite{diehl:2005:real}, which updates the applied control input at every major SQP-iteration. In this work, by contrast, we run the NLP solver until a sufficient degree of convergence is met to update the whole trajectory at once. Replanning starts immediately after the initial problem is solved. In order to ensure a smooth trajectory, the initial state for replanning is interpolated from the last reference trajectory using the estimated replanning time. The convergence of the NLP solver is sped up by initializing it with the previous optimal state- and input trajectories. The previous solution gets interpolated such that the nodes are again equally distributed over the remaining time horizon. 
The optimizer keeps replanning as long as a feasible solution can be found and the robot arm is a minimum distance away from the target, which avoids over-parametrization towards the trajectory end.

As initial time horizon, we select $T= 5.0$~s and discretize it into 20~equally spaced shots. We choose zero-order hold interpolation of the control inputs between the nodes and propagate the system state with a constant RK4 step-size of 50~ms. The initial guess for the state decision variables consists of zero joint velocities and equidistantly spaced joint positions that are obtained by direct interpolation between the initial and terminal joint configuration. The initial guess for the control input decision variables are the steady-state joint torques computed by the Recursive Newton Euler inverse dynamics evaluated at the corresponding states. 

\subsubsection{Results}
In~\cite{neunert:2016:derivatives}, we found that IPOPT significantly outperformed SNOPT in all our simulation experiments (similar results were obtained in~\cite{pardo:2016:evaluating}). We shall anticipate that we found the same to hold for the replanning case. Furthermore, we noticed that SNOPT runtimes varied significantly with the initial and desired arm configuration and the obstacle state while the majority of all recorded IPOPT runtimes lay in a relatively narrow band.
As a consequence, and for reasons of limited space, we only focus on IPOPT in the remainder of this paper.

\begin{figure}[tbp]
    \centering
    \includegraphics[width = 0.99\columnwidth]{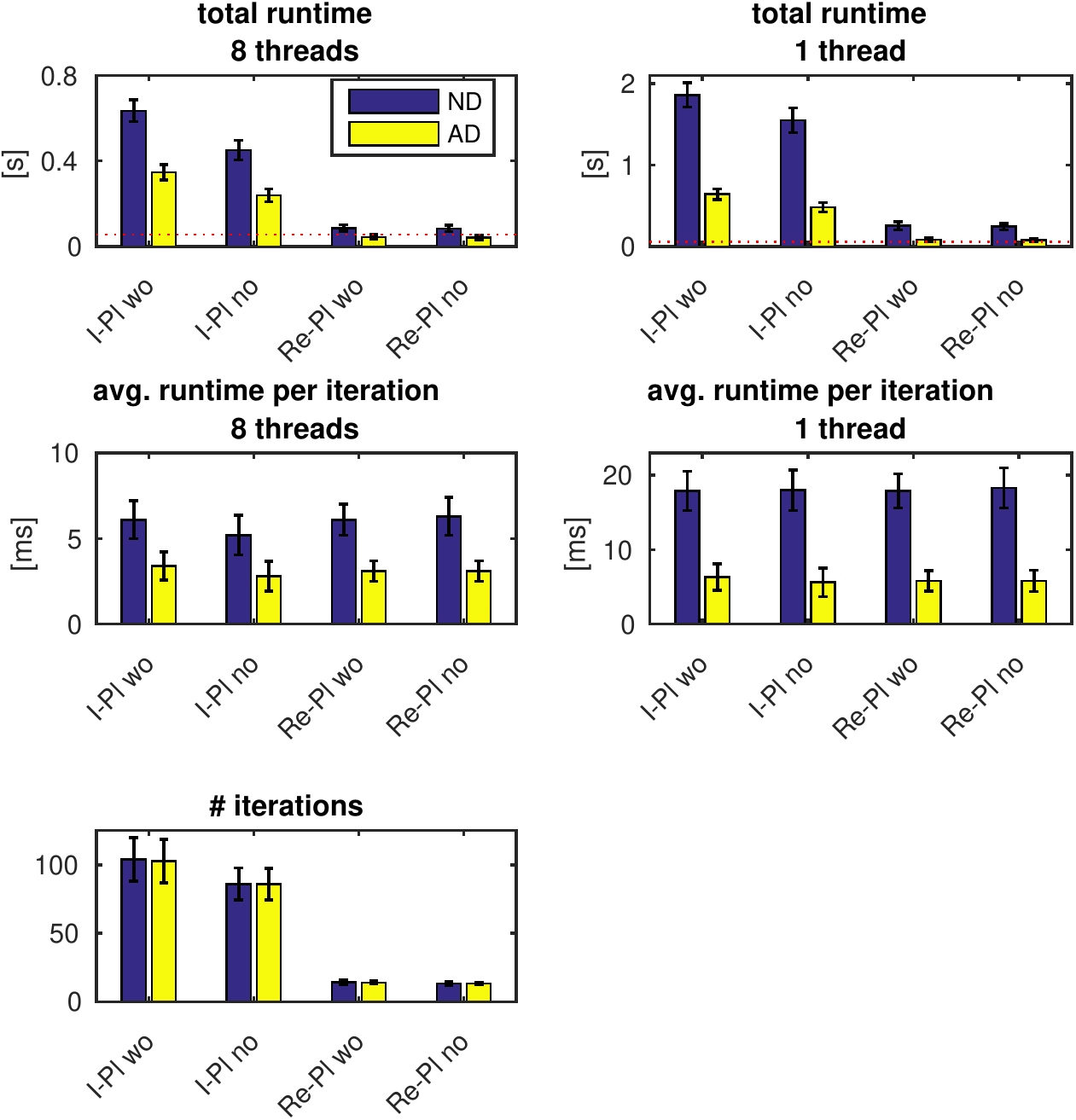}
    \caption{Comparison between Num-Diff (`ND', blue) and Auto-Diff codegen (`AD', yellow) in replanning experiments. We show timings for the initial planning step ('I-Pl') and the replanning ('Re-Pl'), for cases with obstacle ('wo') and without ('no'). These timings were recorded for different numbers of parallel threads (1/8).
    The plot shows average data and standard deviations from 20 experiments.
    The dotted, horizontal lines in the two top diagrams indicate the update time of the obstacle state estimator (50~ms).
    Results show that Auto-Diff codegen leads to significantly improved runtime. The relative difference is remarkable in the single-core case. 
    Thanks to Auto-Diff codegen, it is possible to achieve replanning times that are consistently lower than the obstacle estimator's update rate.
    }
    \label{fig:dms_results}
\end{figure}
\begin{figure*}[tb]
\setlength{\tabcolsep}{-1pt}  
\renewcommand{\arraystretch}{0.1}
\begin{tabular}{c}  
\subfloat[The arm avoids a moving obstacle during a go-to task. The obstacle is defined by the fiducial marker mounted on a stick guided by a human.]{
\begin{tabular}{cccccc}  
    \includegraphics[width = 0.163\textwidth]{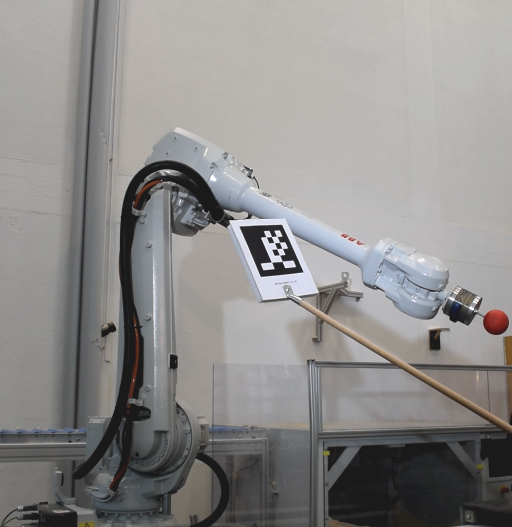} \hfill
&   \includegraphics[width = 0.163\textwidth]{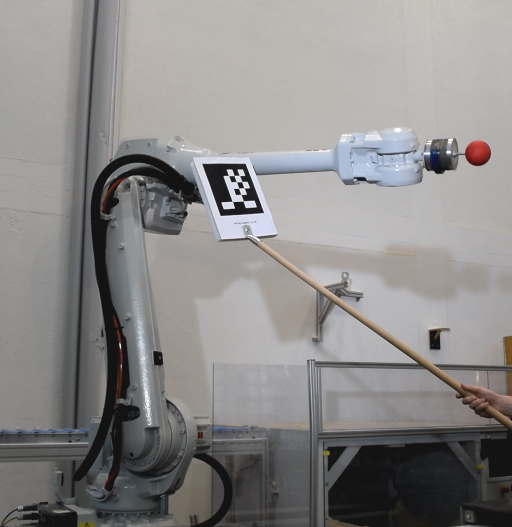} \hfill
&   \includegraphics[width = 0.163\textwidth]{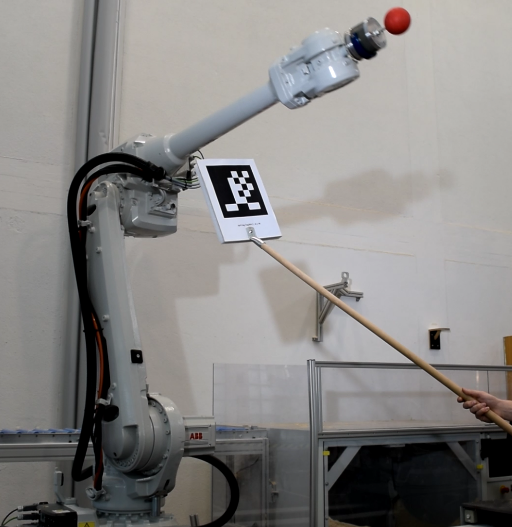} \hfill
&   \includegraphics[width = 0.163\textwidth]{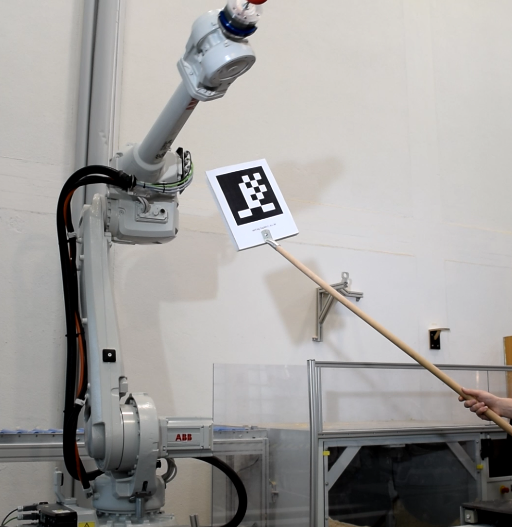} \hfill
&   \includegraphics[width = 0.163\textwidth]{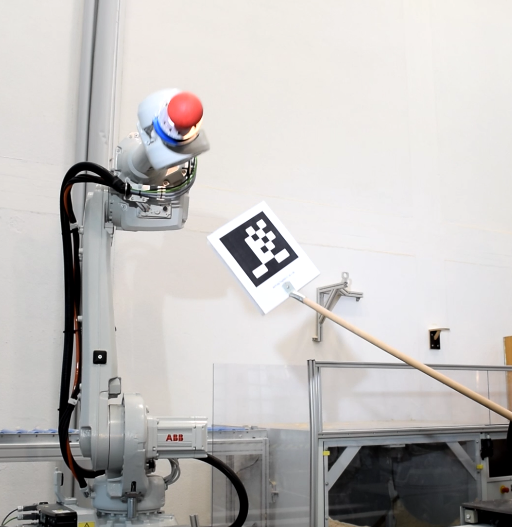} \hfill
&   \includegraphics[width = 0.163\textwidth]{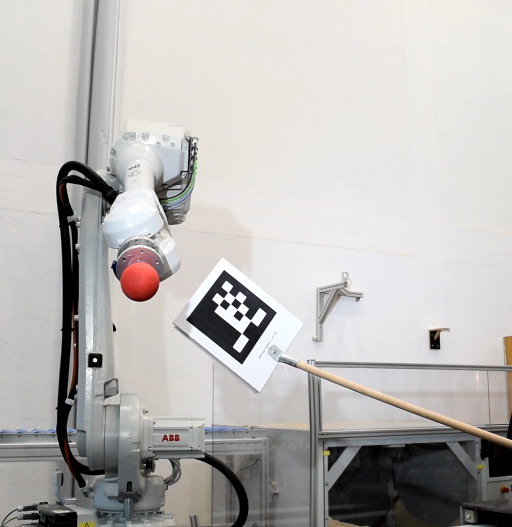} \hfill
\end{tabular}
\label{fig:abb_robot_fig} 
}
\hfill
\vspace*{-0.7em} 
\\
\subfloat[Visualizations of the arm, the obstacle (orange sphere) and the end-effector trajectories. The latter illustrate how the optimal plan changes over the replanning cycles.]{
\begin{tabular}{cccccc}  
    \includegraphics[width = 0.163\textwidth]{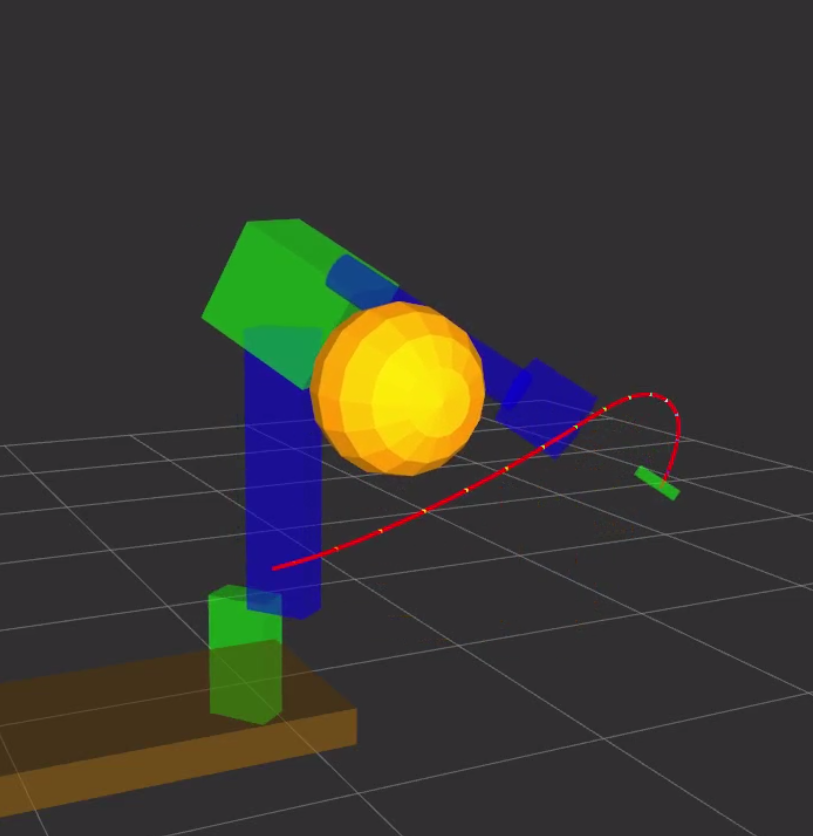} \hfill
&   \includegraphics[width = 0.163\textwidth]{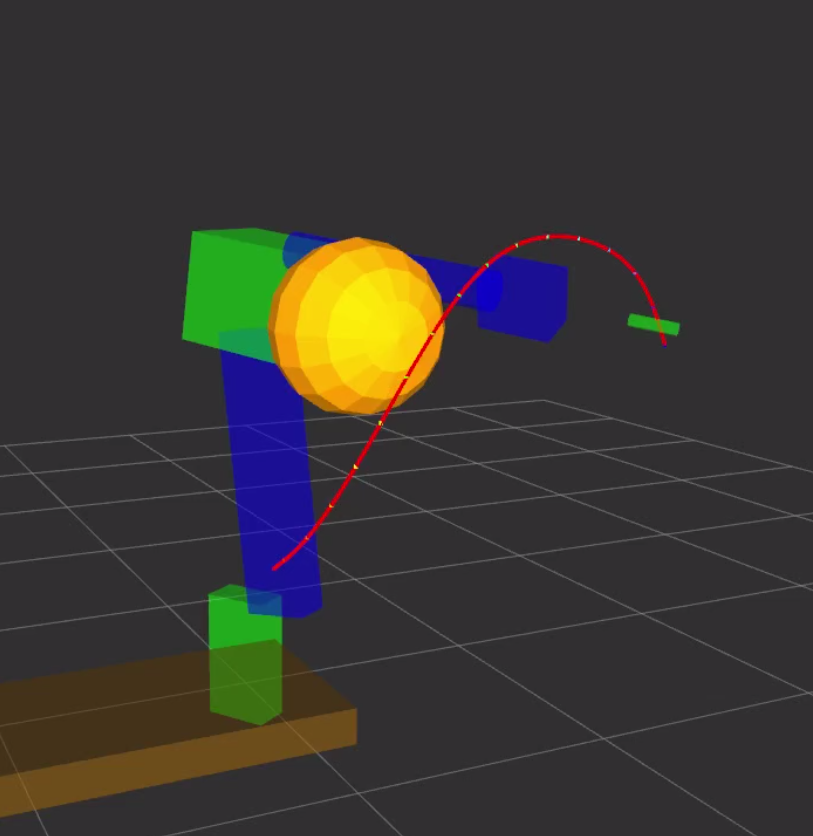} \hfill
&   \includegraphics[width = 0.163\textwidth]{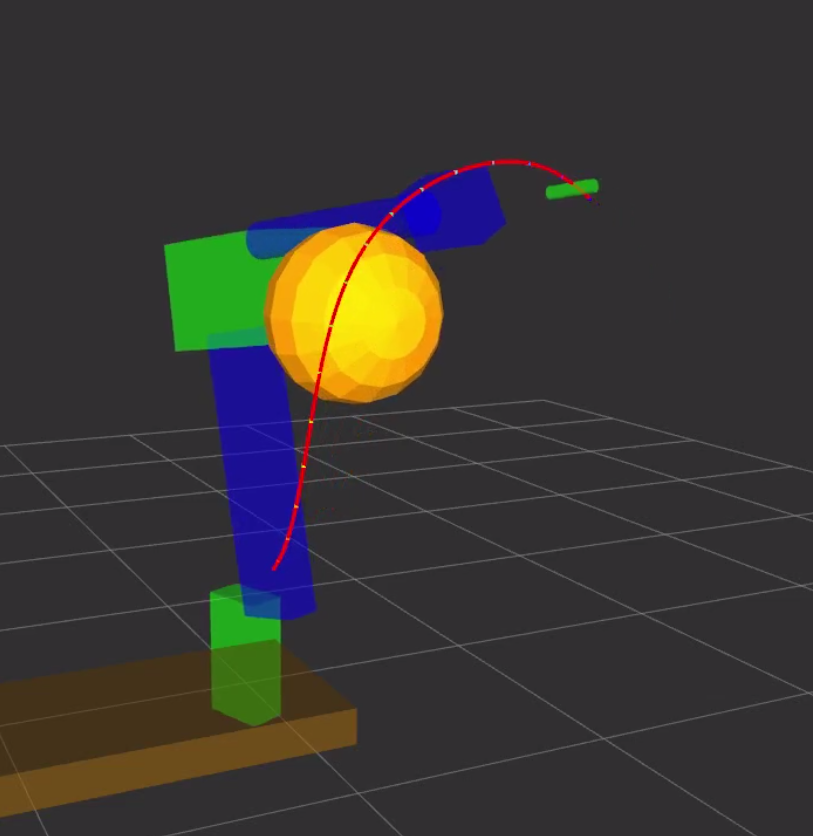} \hfill
&   \includegraphics[width = 0.163\textwidth]{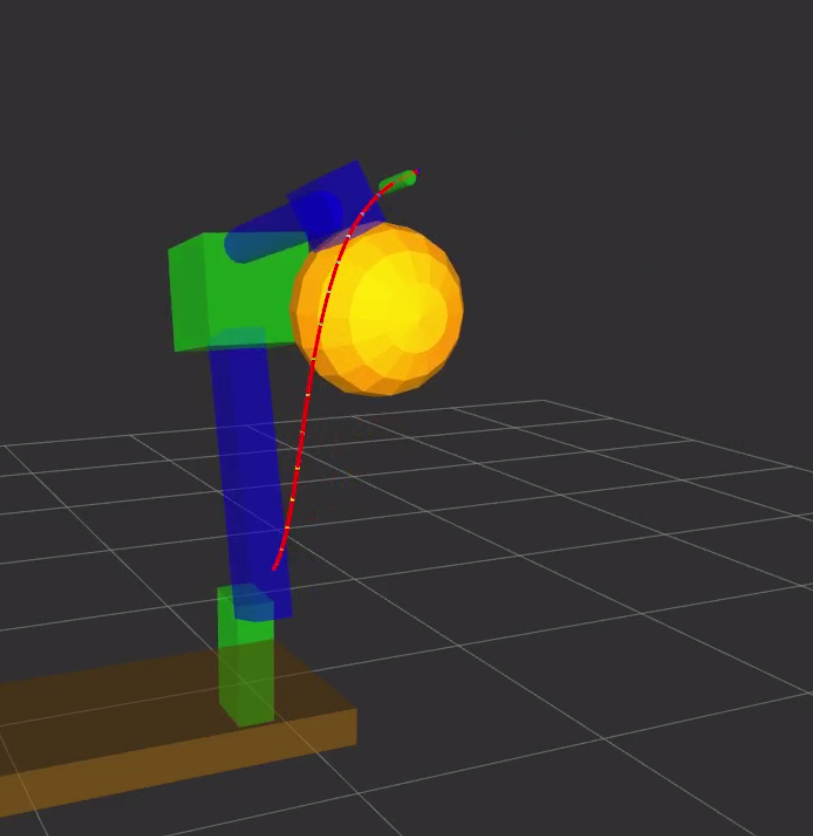} \hfill
&   \includegraphics[width = 0.163\textwidth]{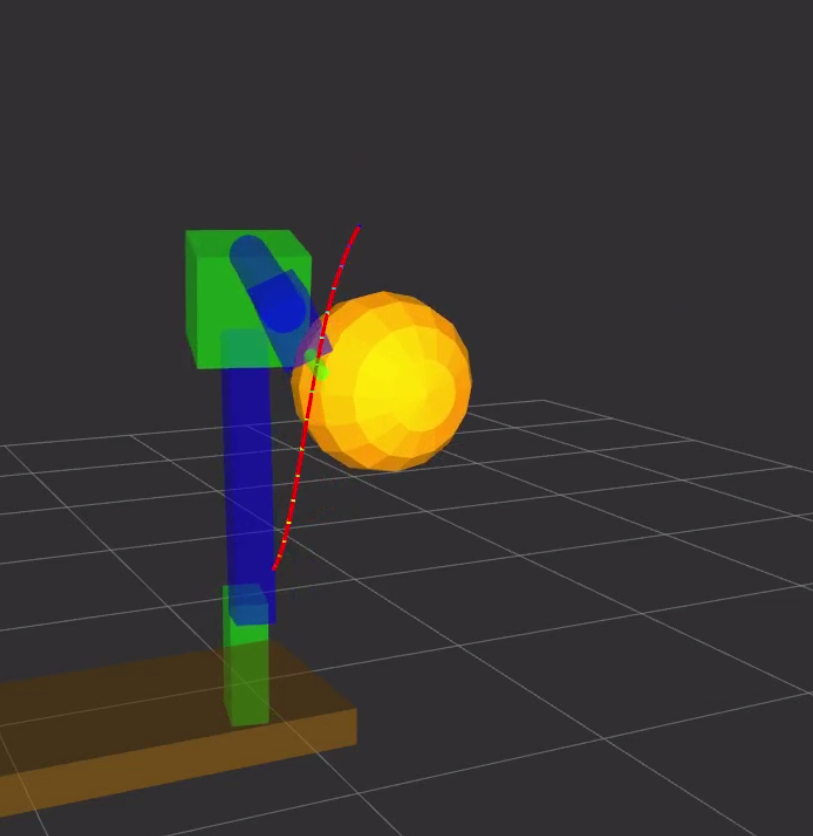} \hfill
&   \includegraphics[width = 0.163\textwidth]{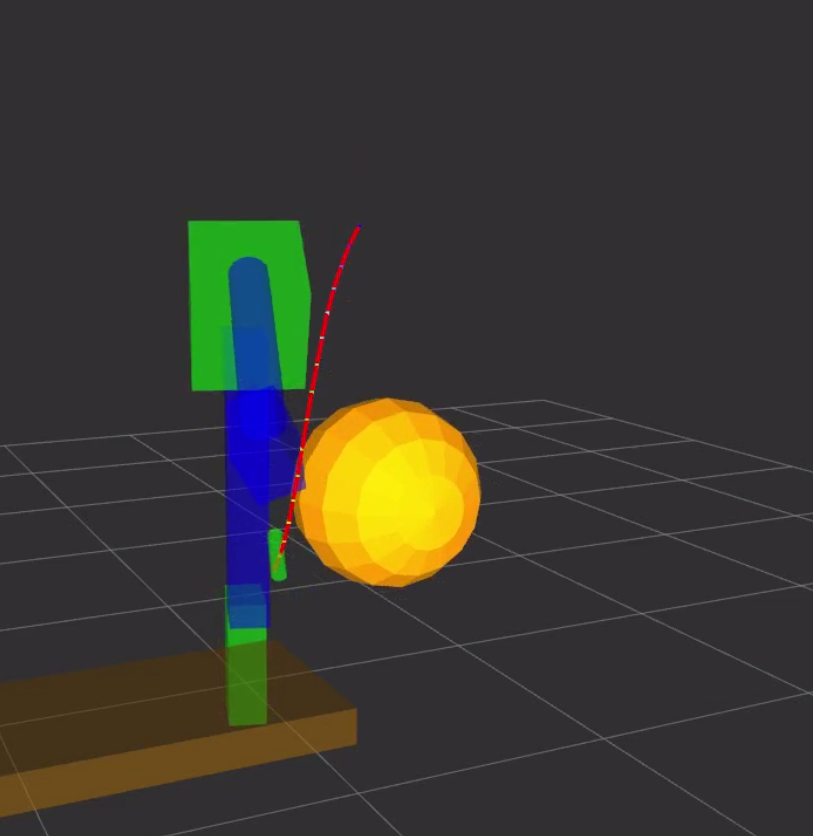} \hfill
\end{tabular}
\label{fig:abb_rviz_fig} 
}
\hfill 
\end{tabular} 
\renewcommand{\arraystretch}{1.0}
\caption{An image sequence of replanning experiments on the ABB IRB4600 arm (a) and corresponding visualizations of the obstacle and the trajectories (b). All images taken equidistant in time with $\Delta t= 1$~s.}
\label{fig:hardware_experiments}
\end{figure*}

In the following, we highlight the performance gain for DMS with replanning when using Auto-Diff generated derivatives of Rigid Body Dynamics instead of numerical derivatives. To assess the performance, we look at the runtime and the number of iterations that IPOPT requires for different scenarios. We investigated 20 planning and replanning scenarios with moderately varied joint positions and different obstacle movements.
For showing the influence of the collision constraints on the runtime, we repeated all experiments without obstacle and provide the recorded timings as reference.
Figure~\ref{fig:dms_results} summarizes the results, where the planner was run on an Intel Xeon E5 processor. It is evident that independent of the scenario or planning stage, Auto-Diff codegen outperforms Num-Diff in runtime by a factor of usually 100\% or more. 
During replanning, we reach an average of 13-15 IPOPT iterations. In the multi-threaded setup with Auto-Diff codegen derivatives, this leads to replanning times reliably lower than the update rate of the vision system (50~ms).
This allows us to optimally exploit the obstacle estimator information by running the replanner synchronously with the vision system and thus obtain optimal reactiveness and replanning performance.
Figure~\ref{fig:hardware_experiments} shows an image sequence from one of the hardware experiments. It gives an example of how the optimal trajectory gets updated as the obstacle is moved in the robot's workspace. 

\section{Discussion and Outlook}
In this work, we propose how Auto-Diff can be efficiently applied to RBD algorithms. Based on this study we extended our open source Robotics Code Generator, enabling full Automatic Differentiation compatibility. This allows us to underline the importance of generating source code for the derivatives obtained from Auto-Diff, avoiding the usual overhead during evaluation. The resulting derivative code outperforms even carefully hand-designed and optimized analytical derivative implementations.
Besides basic timings of forward and inverse dynamics derivatives, we show that Auto-Diff with generated code can significantly improve performance in control and optimization applications.

In the motion planning examples for the legged robot HyQ, Auto-Diff codegen allowed us to lower the runtime of our TO from minutes to seconds, achieving an overall speedup of 500\%. By auto-differentiating the entire dynamics including the contact model, we are able to achieve fast but also accurate derivatives without manually deriving them. 
While this work is mostly focused on dynamics and kinematics, the approach naturally extends to arbitrary constraints, as for example shown in~\cite{giftthaler2017efficient} for complex non-holonomic constraints.
Therefore, we believe that Auto-Diff will play an important role in running optimal control algorithms online as Model Predictive Controllers.

In our hardware experiments on a 6~DoF robot arm, Auto-Diff codegen allowed us to achieve replanning rates fast enough to be synchronized with a typical robotic vision system. In this example, Auto-Diff provided the critical speedup to convert our optimization based motion planner to an online continuous replanning approach.

While the examples in this paper are still partially academic, they underline the potential of Auto-Diff when dealing with RBD. This motivates us to include Auto-Diff codegen in our more advanced controllers and estimators. In this work, we have already seen a significant gain from generating first order derivatives. However, although not detailed in this work, our tool chain allows us to generate second order derivatives as well. This can further speed up our control approaches which currently only approximate second order derivatives.

\section*{Source Code and Examples}
\footnotesize{
The latest version of the code generator `RobCoGen' is available at \mbox{\url{https://robcogenteam.bitbucket.io}}. As an example, the Auto-Diff compatible RobCoGen output for the quadruped HyQ as well as all derivative timing and accuracy tests are available at~\cite{neunert:2017:controltoolbox} and \url{https://bitbucket.org/adrlab/hyq_gen_ad}.
}

\section*{Funding}
\footnotesize{
This research was supported by the Swiss National Science Foundation through the National Centre of Competence in Research (NCCR) Robotics, the NCCR Digital Fabrication, and a Professorship Award to Jonas Buchli. 
}


\bibliographystyle{ieeetr}
\bibliography{root}

\end{document}